\pdfoutput = 1
\documentclass[twocolumn]{article}

\usepackage{times}
\usepackage[numbers]{natbib}
\usepackage[english]{babel}
\usepackage{blindtext}
\usepackage{graphicx}
\usepackage{amsmath, amsthm, amssymb, bbm, bm}
\usepackage{cases}
\usepackage{enumerate}
\usepackage{float}      
\usepackage{subcaption}  
\usepackage{wrapfig}
\usepackage[margin=0cm]{caption}
\usepackage[titletoc, toc]{appendix}
\usepackage{tabularx}

\usepackage{multirow}
\usepackage{makecell}
\usepackage{placeins}  

\usepackage[x11names, usenames, dvipsnames, svgnames, table]{xcolor}
\definecolor{firebrick}{rgb}{.698,.133,.133}
\definecolor{mybluelight}{rgb}{0.9, 0.9, 1.}

\usepackage[utf8]{inputenc} 
\usepackage[T1]{fontenc}    
\usepackage{url}            
\usepackage{booktabs, colortbl}       
\usepackage{amsfonts}       
\usepackage{nicefrac}       
\usepackage{microtype}      

\usepackage{csquotes}
\usepackage{latexsym}

\usepackage[boxruled, vlined, linesnumbered]{algorithm2e}
\SetAlFnt{\small}
\SetAlCapFnt{\small}
\SetAlCapNameFnt{\small}
\usepackage{algorithmic}
\algsetup{linenosize=\tiny}

\let\oldnl\nl
\newcommand{\nonl}{\renewcommand{\nl}{\let\nl\oldnl}}

\usepackage{paralist}

\usepackage{xspace}
\usepackage{soul}
\usepackage{dsfont}
\usepackage{stmaryrd}
\usepackage[textwidth=15mm]{todonotes}
\usepackage{dirtytalk}
\usepackage{pbox}
\usepackage{cprotect}

\usepackage{verbatim}
\usepackage{textcomp}
\usepackage[normalem]{ulem}

\usepackage{mathtools}
\usepackage{etextools}
\usepackage[inline]{enumitem}

\usepackage[colorlinks=true,allcolors=firebrick,bookmarks=false]{hyperref}

\let\OLDthebibliography\thebibliography
\renewcommand\thebibliography[1]{
  \OLDthebibliography{#1}
  \setlength{\parskip}{0pt}
  \setlength{\itemsep}{0pt plus 0.3ex}
}

\newcommand{\kl}{\mathbf{KL}\divx}
\newcommand{\ceq}{\stackrel{\mathclap{\normalfont\mbox{c}}}{=}}

\DeclareMathOperator*{\argmax}{argmax}

\theoremstyle{definition}

\DeclarePairedDelimiterX{\divx}[2]{(}{)}{%
  #1\;\delimsize\|\;#2%
}

\newcommand*{\ie}{\emph{i.e.}\@\xspace}

\usepackage{style}

\makeatletter
\@namedef{ver@everyshi.sty}{}
\newcommand{\removelatexerror}{\let\@latex@error\@gobble}

\title{Deep Interpretable Classification and Weakly-Supervised \\
 Segmentation of Histology Images via Max-Min Uncertainty}

\renewcommand\footnotemark{}

\author{Soufiane~Belharbi$^{1}$,
  ~Jérôme Rony$^{1}$,
  ~Jose Dolz$^{2}$,
  ~Ismail~Ben~Ayed$^{1}$,
  ~Luke~McCaffrey$^{3}$, and
  ~Eric~Granger$^{1}$\\
 	$^1$ LIVIA, Dept. of Systems Engineering, École de technologie supérieure, Montreal, Canada \\
  $^2$ LIVIA, Dept. of Computer Engineering, École de technologie supérieure, Montreal, Canada\\
	$^3$ Goodman Cancer Research Centre, Dept. of Oncology, McGill University, Montreal, Canada\\
{\tt\footnotesize \textcolor{black}{soufiane.belharbi.1@ens.etsmtl.ca, \{jose.dolz, ismail.benayed, eric.granger\}@etsmtl.ca, } }\\
{\tt\footnotesize \textcolor{black}{jerome.rony.1@etsmtl.net, luke.mccaffrey@mcgill.ca}}
\\
}

\newcommand{\ignore}[1]{}

\begin{document}

\maketitle\thispagestyle{fancy}

\begin{abstract}
  Weakly-supervised learning (WSL) has recently triggered substantial interest as it mitigates the lack of pixel-wise annotations.
  Given global image labels, WSL methods yield pixel-level predictions (segmentations), which enable to interpret class predictions. Despite their recent success, mostly with natural images, such methods can face important challenges when the foreground and background regions have similar visual cues, yielding high false-positive rates in segmentations, as is the case in challenging histology images. WSL training is commonly driven by standard classification losses, which implicitly maximize model confidence, and locate the discriminative regions linked to classification decisions. Therefore, they lack mechanisms for modeling explicitly non-discriminative regions and reducing false-positive rates. We propose novel regularization terms, which enable the model to seek both non-discriminative and discriminative regions, while discouraging unbalanced segmentations. We introduce high uncertainty as a criterion to localize non-discriminative regions that do not affect classifier decision, and describe it with original Kullback-Leibler (KL) divergence losses evaluating the deviation of posterior predictions from the uniform distribution. Our KL terms encourage high uncertainty of the model when the latter inputs the latent non-discriminative regions. Our loss integrates: (i) a cross-entropy seeking a foreground, where model confidence about class prediction is high; (ii) a KL regularizer seeking a background, where model uncertainty is high; and (iii) log-barrier terms discouraging unbalanced segmentations. Comprehensive experiments and ablation studies over the public GlaS colon cancer data and a Camelyon16 patch-based benchmark for breast cancer show substantial improvements over state-of-the-art WSL methods, and confirm the effect of our new regularizers. Our code is publicly available\footnote{Code: \href{https://github.com/sbelharbi/deep-wsl-histo-min-max-uncertainty}{{\color{blue} https://github.com/sbelharbi/deep-wsl-histo-min-max-uncertainty}}.}.
\end{abstract}


%
%

\section{Introduction}
\label{sec:introduction}
\begin{figure}[h!]
  \center
\includegraphics[width=\linewidth]{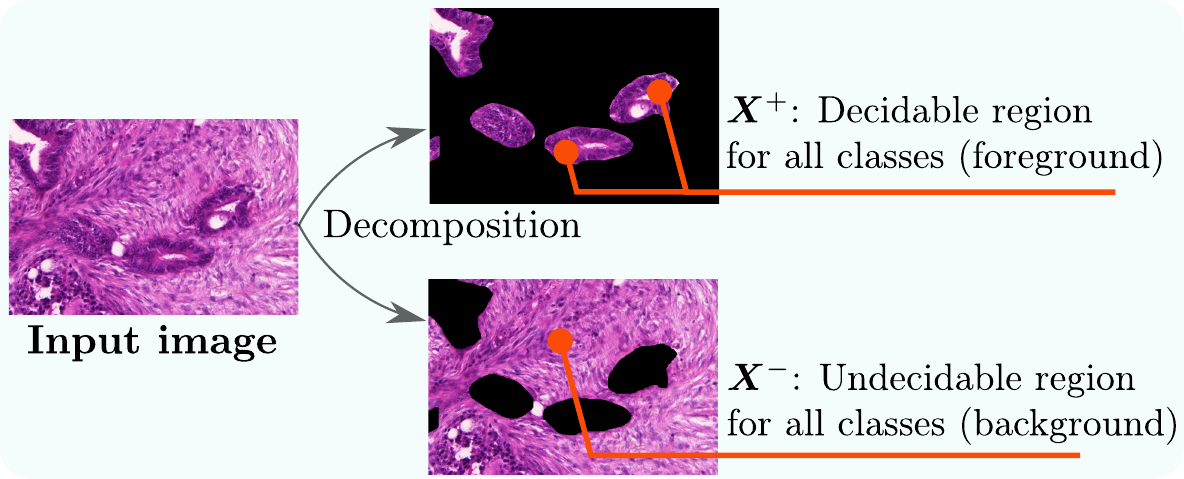}\\
\vspace{2mm}
\includegraphics[width=1.\linewidth]{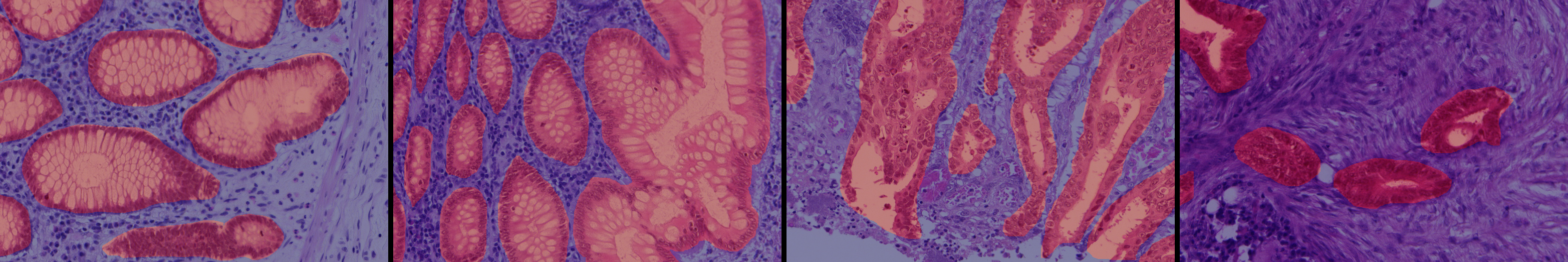}
\caption{\textbf{Top row}: Intuition of our proposal. A decidable region (high confidence) covers the discriminative parts (foreground), while an undecidable region (high uncertainty) covers the non-discriminative parts (background). \textbf{Bottom row}: Examples of test image samples from different classes of the GlaS dataset \cite{sirinukunwattana2017gland}, where the annotated glands are the regions of interest, and the remaining tissue is background. Note the glands' different shapes, sizes, context, and multiple-instance aspects. Best visualized with color.}
\label{fig:fig-0}
\end{figure}

Medical imaging is one of the primary tools for early detection of cancer. In particular, the analysis of histology images remains the gold standard in the assessment of many pathologies, such as breast \cite{he2012histology,gurcan2009histopathological,Veta2014}, colon \cite{Shapcott2019,sirinukunwattana2015stochastic,xu2020colorectal} and brain cancer \cite{ker2019,khalsa2020automated,xu2017large}.
Such analysis is mainly performed manually by pathologists on large histology images. To alleviate the workload of pathologists, computer-aided diagnosis (CAD) has been largely explored to support timely and reliable decisions. In histology images, CAD often relies on computer vision and machine learning algorithms, with a recent focus on deep learning models \cite{rony2019weak-loc-histo-survey}, where image classification has attracted much of the attention \cite{campanella2019clinical,KOMURA201834,rony2019weak-loc-histo-survey,spanhol-breakhis2016,Spanhol2016,SUDHARSHAN2019103}.

In the recent years, deep learning models, and in particular convolutional neural networks, have achieved state-of-the-art performances in a breadth of computer vision and medical imaging problems, for instance, image classification \cite{Goodfellow-et-al-2016,krizhevsky12} and semantic segmentation \cite{dolz20183d,LITJENS201760,LongSDcvpr15}, among many other problems. Despite their unprecedented success in recent years, training highly accurate deep learning models typically requires large annotated data. While building image-level annotations by human experts could be manageable, producing pixel-level annotations is a very laborious and time-consuming task, more so when dealing with histology images, where the very large sizes of the images make manual segmentation intractable in clinical practices.

Despite its intrinsic challenges \cite{choe2020evaluating}, weakly-supervised learning (WSL) has recently emerged as an alternative to reduce the cost and burden of fully annotating large data sets \cite{zhou2017brief}. Models trained through WSL exploit unlabeled inputs, as well as coarse (group-level) or ambiguous labels, which alleviate the need for dense labels. In segmentation, weak supervision could take different forms, including scribbles \cite{Lin2016,ncloss:cvpr18}, points \cite{Bearman2016}, bounding boxes \cite{dai2015boxsup,Khoreva2017,kervadec2020bounding}, global image statistics such as the target-region size \cite{kervadec2019constrained,bateson2019constrained,jia2017constrained,kervadec2019curriculum}, and image-level labels \cite{belharbi2022fcam,kim2017two,pathak2015constrained,teh2016attention,wei2017object}. This paper focuses on the latter case -- WSL models trained with global image-level class annotations. The goal is to classify histology images with only global image-level labels, while producing pixel-level label predictions, thereby localizing the important regions of interest that are linked to the model’s global decision. Pinpointing image sub-regions that were used by the model to make its global image-class prediction not only provides weakly-supervised segmentation, but also enables interpretable deep-network classifiers. It is worth noting that such interpretability aspects are also attracting wide interest in computer vision \cite{bach2015pixel,bau2017network,bhatt2020evaluating,dabkowski2017real,Escalante2018,fong2019understanding,Fong2017ICCV,goh2020understanding,osman2020towards,Murdoch22071,Petsiuk2020,PetsiukDS18,Ribeiro2016kdd,samek2020toward,samek2017explainable,zhang2020interpretable,belharbi2020DeepAlJoinClSegWeakAnn} and medical imaging
\cite{delatorre2020deep,gondal2017weakly,GonzalezGonzalo2020,Taly2019,QUELLEC2017178,keel2019visualizing,Wang2017zoomin}.

Deep learning classifiers are often considered as ``black boxes'' due to the lack of explanatory factors in their decisions. Therefore, the development of learning models that provide such explanatory factors is of capital importance, more so when dealing with sensitive applications as found in medical image analysis, where the transparency issue (i.e., the absence of clear explanatory factors of a model's decision) is a potential liability for machine learning models.

State-of-the-art deep WSL models trained using image-class labels rely heavily on pixel-wise activation maps, often referred to as class activation maps (CAMs) \cite{rony2019weak-loc-histo-survey,Zhang2018VisualInterp,zhou2016learning}. Typically, activation maps are first generated using a classification network, highlighting the relevant image sub-regions that are responsible for the prediction of a given class. Then, these maps may be employed as ``fake'' labels to train segmentation networks, thereby mimicking full supervision. Existing WSL methods could be categorized along two main veins \cite{rony2019weak-loc-histo-survey}: (1) {\em bottom-up} methods, which rely on the input signal to locate the regions of interest, e.g., spatial pooling techniques over activation maps \cite{durand2017wildcat,oquab2015object,sun2016pronet,zhang2018adversarial,zhou2016learning}, multi-instance learning \cite{ilse2018attention} and attend-and-erase based methods \cite{kim2017two,LiWPE018CVPR,SinghL17,wei2017object}; and (2)
{\em top-down} methods, which are inspired by human visual attention.
Initially conceived as visual explanatory tools \cite{Simonyan14a,Springenberg15a,zeiler2014ECCV}, these methods have gained in popularity in weakly-supervised segmentation. They rely on both the input signal and a selective backward signal to determine the region of interest, e.g., special feedback layers \cite{cao2015look}, back-propagation error \cite{zhang2018top} and Grad-CAM \cite{ChattopadhyaySH18wacv,selvaraju2017grad}.

Current WSL methods have yielded promising performances in the context of natural images where, typically, target regions have color distributions that are significantly different from the background. However, weakly-supervised segmentation of histology images faces accrued challenges, as the target regions have image appearances similar to those of the background, and intra-class variations are high across samples, making WSL difficult. These difficulties limit the direct application of the existing deep WSL models, which were mostly designed for and evaluated on natural images. As shown in the recent experimental study in \cite{rony2019weak-loc-histo-survey}, existing WSL methods may face serious challenges when dealing with ambiguous histology images, typically yielding high false-positive rates in pixel-wise predictions. Driven by standard supervised classification losses
(e.g., cross-entropy), these methods implicitly maximize model confidence, and localize the discriminative image sub-regions that have driven the classification decision. Therefore, they lack mechanisms for explicit modeling of non-discriminative regions and reducing false-positive rates, yielding unsatisfying masks. Thus, in order to accurately find the foreground/background regions, the models require additional regularization, accounting for both discriminative and non-discriminative regions.

\textbf{Our main contributions are summarized below}:
\begin{itemize}[leftmargin=*]
     \item We propose novel regularization terms for the weakly-supervised segmentation scenario, which explicitly constrain the model to seek both non-discriminative and discriminative regions, while discouraging unbalanced segmentations. In particular, we leverage the uncertainty of the predictions to identify and constrain irrelevant regions that do not impact the classifier outcomes. These regions lack semantic information required by the model to make decisions, which is translated into high uncertainty areas.
     \item To describe such a high uncertainty, we resort to the Kullback-Leibler (KL) divergence, which evaluates the deviation of posterior predictions from the uniform distribution. Combined with standard cross-entropy, which seeks most-certain regions (\ie foreground), minimizing the joint learning objective pushes the model to explicitly search for background regions.
     \item Furthermore, we consider an additional regularization term expressed with region sizes. Based on log-barrier methods, this term penalizes disproportionate foreground and background regions without extra supervision.
     \item To evaluate our method, we report comprehensive experiments on the public GlaS colon cancer data set and Camelyon16 patch-based benchmark for breast cancer, which demonstrate substantial improvements over state-of-the-art WSL methods. In addition, we report ablation studies on different components of our method, which confirm that the observed improvements are indeed due to the novel uncertainty regularizer and regional size constraints.
 \end{itemize}

\begin{figure*}[h!]
  \centering
\includegraphics[width=0.7\linewidth]{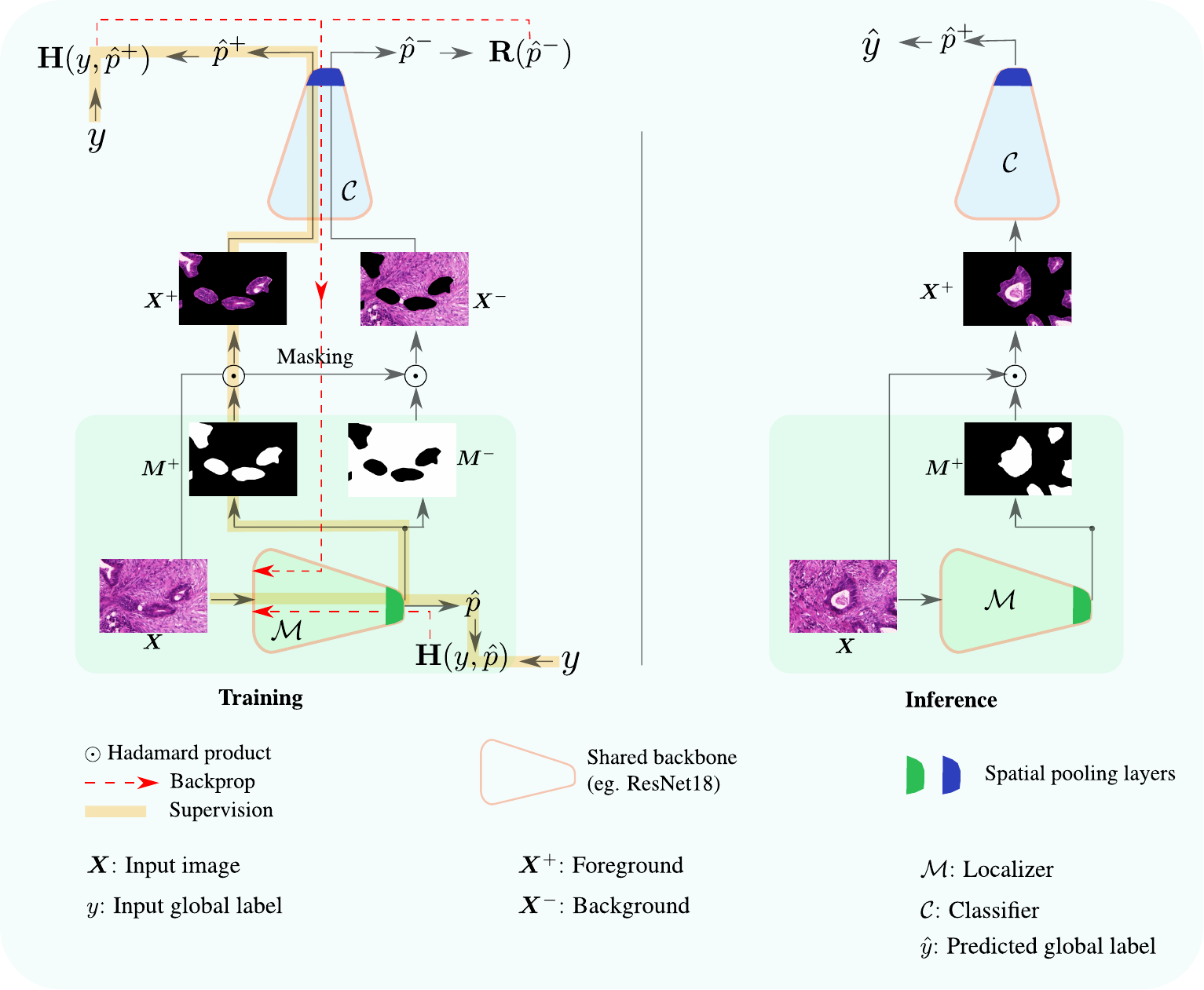}
\caption{Illustration of our approach. Training phase (left): Image ${\bm{X}}$ is presented to localizer ${\mathcal{M}}$, which computes the foreground and background masks. The masked input images are then fed to classifier ${\mathcal{C}}$. The classifier is encouraged to be certain about the class of foreground image ${\bm{X}^+}$, while being uncertain for background image ${\bm{X}^-}$. This training is performed using only the global image-class labels. Inference phase (right):  For input ${\bm{X}}$, a single forward pass is performed into the localizer and classifier, so as to predict the foreground mask and class label of the image.}
\label{fig:fig-proposal}
\end{figure*}


\section{Max-Min uncertainty framework}
\label{sec:proposedmethod}

Consider a set of training samples ${\mathbb{D} = \{(\bm{X}_i, y_i)\}_{i=1}^n}$, where ${\bm{X}_i}$ is an input image and ${y_i} \in \{1, \cdots, c\}$ its corresponding label (or class). For a given image $\bm{X} \in \{\bm{X}_i\}_{i=1}^n$ with label $y \in \{1, \cdots, c\}$, we define a foreground ${\bm{X}^+}$ as an image sub-region, which provides enough information to predict $y$. In contrast, a background ${\bm{X}^-}$ is a sub-region, which does not hold any relevant information that supports the image class (Fig.\ref{fig:fig-0}). The union of ${\bm{X}^+}$ and ${\bm{X}^-}$ yields ${\bm{X}}$. These sub-images can be represented with binary pixel-wise masks, $\bm{M}^+$ and $\bm{M}^-$, via the Hadamard product:
\[{\bm{X}^+} = {\bm{X} \odot \bm{M}^{+}} \mbox{~and~} {\bm{X}^-} = {\bm{X} \odot \bm{M}^{-}}\]

Consider a deep network classifier with parameters ${\bm{\theta}_{\mathcal{C}}}$, and let posterior $\hat{p} \in [0, 1]^C$ denotes its softmax probability ouput vector\footnote{For the sake of simplicity, we omit the dependence of the network softmax probability outputs on the input images and on the trainable parameters, as this does not yield any ambiguity.} for a given training image $\bm{X}$. Let ${p} \in \{0, 1\}^C$ denotes the one-hot vector encoding of the ground-truth label of $\bm{X}$, i.e., one of the components of $p$ takes value $1$ for the true label while the other components take value $0$. The standard cross-entropy loss for image classification is given by:
\begin{equation}
     \label{eq:eq-0}
     \mathbf{H}(p, \hat{p}) = - p^t \log \hat{p} = - \log \mbox{Pr}(y | \bm{X}),
\end{equation}
where super-script $t$ denotes the transpose operator and $\mbox{Pr}(y | \bm{X})$ is the softmax network output for ground-truth label $y$ and input $\bm{X}$. Notice that ${\mathbf{H}(\hat{p}, \hat{p})}$ is the {\em Shannon entropy}
of distribution ${\hat{p}}$. For simplicity in what follows, we use the following notation to denote the Shannon entropy:
\[{\mathbf{H}(\hat{p})} = \mathbf{H}(\hat{p}, \hat{p}) = - {\hat{p}}^t \log \hat{p}. \]

Standard WSL methods aim at finding \emph{implicitly} ${\bm{M}^+}$, i.e., the regions in the image where the model is {\em confident} about its image-class prediction. Typically, this is achieved by minimizing a cross-entropy loss:
\begin{equation}
     \label{eq:eq-1}
    \min_{\bm{\theta}_{\mathcal{C}}} \quad \mathbf{H}(p, \hat{p}).
\end{equation}
Optimizing \eqref{eq:eq-1} solely may yield unsatisfying masks, either in the form of small discriminative regions, missing significant parts of the target foreground regions \cite{zhou2016learning}, or by considering the entire image as a discriminative foreground (i.e., the background region is empty) \cite{rony2019weak-loc-histo-survey}. Thus, in order to accurately find the foreground/background regions, the model needs additional regularization, accounting for both discriminative and non-discriminative regions. In this work, we explicitly constrain the model to seek both discriminative and non-discriminative regions by optimizing a loss containing two competing terms: (i) a standard cross-entropy loss seeking the largest foreground ${\bm{M}^+}$, where model {\em confidence} about the class prediction is high; and (ii) an original Kullback–Leibler (KL) regularizer seeking the largest background ${\bm{M}^-}$, where the model {\em uncertainty} is high (or confidence is low). Given background input ${\bm{X}^-}$, our KL regularizer encourages the model predictions to match a {\em uniform} distribution, which corresponds to maximum  uncertainty (or confusion) since  all  the  classes  are equi-probable.

\textbf{(a) Foreground localization ${\bm{X}^+}$:} The first term in our model is a standard cross-entropy as in \eqref{eq:eq-1}, but with the input image corresponding to foreground localization ${\bm{X}^+}$:
\begin{equation}
     \label{eq:eq-1-1}
     \min_{\bm{\theta}_{\mathcal{C}}} \quad \mathbf{H}(p, \hat{p}^+),
\end{equation}
where ${\hat{p}^+}$ is the softmax posterior probability vector conditioned over input image ${\bm{X}^+}$ (Fig. \ref{fig:fig-proposal}). It is well-known that minimizing the cross-entropy in \eqref{eq:eq-1-1}
encourages posterior predictions $\hat{p}^+$ to be close to the vertices of the simplex in $\{0, 1\}^C$ (i.e., $\hat{p}^+$ approaches binary posterior predictions) and, hence, to be confident. This means that the
softmax posterior prediction for one label is close to $1$ while all the other predictions are close to $0$. This can be seen immediately from the fact that ground-truth vector $p$ is binary and the minimum of
\eqref{eq:eq-1-1} is achieved when prediction $\hat{p}^+$ matches exactly $p$. Therefore, minimizing \eqref{eq:eq-1-1} encourages high confidence of the model when the latter takes the foreground region
as input.

\textbf{(b) Background localization ${\bm{X}^-}$:} We define the background region as the part of the image, where the model is most uncertain about class predictions, due to the lack of evidence to support any
of the classes. We describe high uncertainty of the model with a KL divergence evaluating the deviation of posterior predictions from the uniform distribution (i.e., deviation from the middle of the simplex, which corresponds to a maximum amount of uncertainty). Therefore, minimizing such a KL divergence encourages high uncertainty (or low confidence) of the model when the latter takes the background region ${\bm{X}^-}$ as input. We consider using high uncertainty over the background as a \emph{characteristic} to localize irrelevant parts in the image (i.e., the parts that do not affect model decision as to the class of the image).
The intuition behind this is that our KL regularizer should increase the awareness of the model as to the presence of non-informative regions and reduce the space of possible masks obtained from minimizing the cross-entropy
in \eqref{eq:eq-1-1} alone (e.g., the trivial solution where the whole image is considered as a foreground region). As we will see in our experiments through a comprehensive ablation study, our KL regularizer has an important effect on the performances, reducing the amount of false positives. We propose two versions of the KL divergence.

\emph{(1) Explicit Entropy Maximization (EEM):} The first version of our KL regularizer is equivalent (up to an additive constant) to maximizing the entropy of predictions given background input ${\bm{X}^-}$, or to minimizing the negative entropy:
\begin{equation}
     \label{eq:eq-2}
     \min_{\bm{\theta}_{\mathcal{C}}} \quad -\mathbf{H}(\hat{p}^-) \; ,
\end{equation}
where ${\hat{p}^-}$ is the softmax posterior probability vector conditioned over input image ${\bm{X}^-}$ (Fig. \ref{fig:fig-proposal}).
It is straightforward to notice the following:
\begin{align}
     \label{eq:eq-3}
     -\mathbf{H}(\hat{p}^-) &= \kl{\hat{p}^-}{q} - \mathbf{H}(\hat{p}^-, q) \nonumber \\
                     &= \kl{\hat{p}^-}{q} - \log (c)  \nonumber \\
                     &\ceq \kl{\hat{p}^-}{q},
\end{align}
where ${q}$ is the uniform distribution,  ${\kl{\cdot} {\cdot}}$ is the Kullback–Leibler divergence (See Appendix \ref{sec:sec-app-0}), and symbol $\ceq$ denotes equality up to an additive or multiplicative constant.

Our max-uncertainty model in \eqref{eq:eq-2} aims at maximizing the entropy of predictions. Therefore, it should not be confused with entropy minimization, which is widely used in the context of semi-supervised and unsupervised learning \cite{berthelot2019mixmatch,grandvalet2005semi,jabi2019deep,miyato2018virtual}. Entropy minimization encourages high confidence in the predictions, whereas our model in \eqref{eq:eq-2} goes in the opposite direction, promoting high uncertainty (or low confidence) of the posterior predictions given the background input.

\begin{figure}[h!]
  \center
\includegraphics[width=\linewidth]{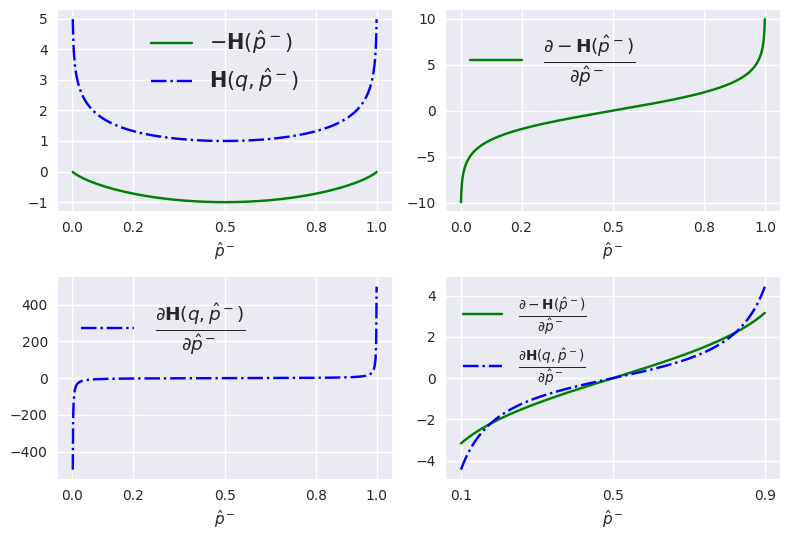}
\caption{Different energies for binary classification and their derivatives (Appendix \ref{sec:sec-app-0}). The bottom-right figure is plotted over the range [0.1, 0.9] in the x-axis (${\hat{p}^-}$). In this interval, notice that the gradient of the entropy is almost linear whereas the gradient of the cross-entropy shows a different dynamic. All the curves are computed using the logarithm to the base of $2$ ($\log_2$).}
\label{fig:fig-energy}
\end{figure}

 \emph{(2) Surrogate for explicit Entropy Maximization (SEM):}
As our goal is to encourage ${\hat{p}^-}$ to be close to the uniform distribution (max-uncertainty), we can also minimize w.r.t network parameters $\bm{\theta}_{\mathcal{C}}$ the following variant of the KL divergence as an alternative to \eqref{eq:eq-3}:
\begin{equation}
     \label{eq:eq-4}
     \kl{q}{\hat{p}^-} \ceq
     \mathbf{H}(q, \hat{p}^-)
\end{equation}
While both \eqref{eq:eq-2} and \eqref{eq:eq-4} are convex with respect to the posterior predictions, and, at the minimum, they both bring ${\hat{p}^-}$ to a uniform distribution (hence, maximum uncertainty/entropy), they exhibit different gradient dynamics. An example for two classes is illustrated in Fig.\ref{fig:fig-energy}.

\textbf{Total training loss}. The total training loss is composed of two terms to localize the foreground and background regions simultaneously. Localizing the background has a regularization effect, reducing false positives without any supervision. To avoid that one region dominates the other, i.e., trivial imbalanced solutions (e.g., the foreground corresponds to the entire image), we constrain the model to find the \emph{largest} foreground/background regions by imposing size constraints formulated through a log-barrier method, which is standard in convex optimization \cite{boyd2004convex}:
\begin{equation}
\label{eq:eq-5}
\begin{aligned}
& \min_{\bm{\theta}_{\mathcal{C}}}
& & \mathbf{H}(p, \hat{p}^+) + \lambda \; \mathbf{R}(\hat{p}^-) - \frac{1}{t} \left[\log\bm{s}^+ + \log\bm{s}^- \right],
\end{aligned}
\end{equation}
where
\begin{numcases}{\mathbf{R}(\hat{p}^-) =}
   - \mathbf{H}(\hat{p}^-) & \text{or} \label{eq:eq-6}
   \\
   \mathbf{H}(q, \hat{p}^-), &  \label{eq:eq-7}
\end{numcases}
${\lambda}$ is a balancing positive scalar, and $t > 0$ is a parameter that determines the accuracy of the approximation of the barrier method. We define the size of each mask as:
\begin{equation}
\label{eq:eq-s+}
    \bm{s}^+ = \sum_{z \in \Omega}\bm{M}^+(z)
\end{equation}
and
\begin{equation}
\label{eq:eq-s-}
    \bm{s}^- = \sum_{z \in \Omega}\bm{M}^-(z),
\end{equation}
where ${\Omega}$ is the spatial image domain.

Note that, when minimizing the log-barrier term alone, the optimal solution is reached when the foreground and background regions have the
same size. However, this perfectly balanced solution does not occur in practice due to the opposing effects of the other classification terms in the overall model, and
to the fact that the log-barrier acts as soft penalty, not as a hard constraint. In fact, there is a competition between the two classification terms over the
foreground and background regions, and these classification terms compete with the log-barrier penalty.
The perfectly balanced solution increases the classification losses and, therefore, is penalized by these losses.
The goal of the log-barrier terms is to penalize trivial, extremely imbalanced solutions, where one region (either the foreground or background) is very large
(almost reaching the image size) while its complement is very small. In the case of ambiguous images such as histology images, weakly-supervised methods tend to activate
over the entire image, yielding foreground masks that entirely dominate the obtained solutions, as will be shown in the next section. The log-barrier terms avoid these
extremely imbalanced solutions. In the experimental section, we will show how using the log-barrier term yields regions sizes that are close to the ground-truth region sizes, not to perfectly balanced solutions.

\textbf{Mask computation:} The mask is learned through another deep learning model (localizer) ${\mathcal{M}(. ;\; \bm{\theta}_{\mathcal{M}})}$, and pseudo-binarized using:
\begin{equation}
    \label{eq:eq-8}
    \bm{M} \coloneqq 1/(1 + \exp(- \omega \times (\bm{M} - \sigma))),
\end{equation}
where ${\omega}$ is a scalar that ensures that the sigmoid approximately equals to $1$ when ${\bm{M}}$ is larger than ${\sigma}$, and approximately equals to $0$ otherwise. To boost the gradient at ${\mathcal{M}}$ and help learning a mask that indicates discriminative regions, ${\mathcal{M}}$ is trained to classify the complete image as well (Fig.\ref{fig:fig-proposal}). We optimize jointly w.r.t parameters ${\bm{\theta}_{\mathcal{M}}}$ and ${\bm{\theta}_{\mathcal{C}}}$ during the stochastic gradient descent optimization process.

\begin{center}
\begin{minipage}{\linewidth}
\IncMargin{0.04in}
\removelatexerror
\begin{algorithm}[H]
    \SetKwInOut{Input}{Input}
     \SetKwInOut{Models}{Models}
    \SetKwInOut{Initial}{Initial}
    \Input{
    ${\mathbf{X}}$: Image, $y$: Global image label.
    }
    \Models{${\mathcal{M}, \mathcal{C}}$}
    \vspace{0.1in}

    \nonl \textrm{// \colorbox{mybluelight}{End-to-end training: one gradient step}} \\
    Forward ${\mathbf{X}}$ into ${\mathcal{M}}$ (Fig.\ref{fig:fig-proposal}). \\
    Compute the mask ${\mathbf{M}}$ (${\mathbf{M}^+,\; \mathbf{M}^-}$)(Eq.\ref{eq:eq-8}). \\
    Compute ${\mathbf{X}^+ = \mathbf{X} \odot \mathbf{M}^+}$, ${\mathbf{X}^- =\mathbf{X} \odot \mathbf{M}^-}$. \\
    Forward ${\mathbf{X}^+,\; \mathbf{X}^-}$ into ${\mathcal{C}}$. \\
    Compute loss in Eq.\ref{eq:eq-5}, and loss of ${\hat{p}}$ (Fig.\ref{fig:fig-proposal}).\\
    Update ${\mathcal{M},\; \mathcal{C}}$ parameters using gradient of both losses.\\
    \nonl \textrm{// \colorbox{mybluelight}{End-to-end evaluation (test):}} \\
    Forward ${\mathbf{X}}$ into ${\mathcal{M}}$ (Fig.\ref{fig:fig-proposal}). \\
    Compute the mask ${\mathbf{M}}$ (${\mathbf{M}^+, \mathbf{M}^-}$)(Eq.\ref{eq:eq-8}). \\
    Compute ${\mathbf{X}^+ = \mathbf{X} \odot \mathbf{M}^+}$. \\
    Forward ${\mathbf{X}^+}$ into ${\mathcal{C}}$. \\
    Output: (a) foreground ${\mathbf{M}^+}$, (b) background ${\mathbf{M}^-}$, (c) Image class: ${\argmax \hat{p}^+}$.
    \vspace{0.1in}
    \caption{Algorithmic description of our method. 
    }
    \label{alg:alg-0}
\end{algorithm}
\DecMargin{0.04in}
\end{minipage}
\end{center}

\section{Experiments}
\label{sec:experiments}

\subsection{Datasets}
Our goal is to evaluate the proposed method in classification and weakly-supervised segmentation tasks. Therefore, histology datasets with both global class- and pixel-level annotations are required.
There are two public datasets that can be employed for validation in our scenario.

\noindent \textbf{(a) GlaS dataset \cite{sirinukunwattana2015stochastic}:} It is a histology dataset for colon cancer diagnosis\footnote{The Gland Segmentation in Colon Histology Images Challenge Contest: \href{https://warwick.ac.uk/fac/sci/dcs/research/tia/glascontest}{https://warwick.ac.uk/fac/sci/dcs/research/tia/glascontest}}. It contains 165 images from 16 Hematoxylin and Eosin (H\&E) histology sections and their corresponding labels. For each image, both pixel-level and image-level annotations for cancer grading (i.e., benign or malign) are provided. The whole dataset is split into training (67 samples), validation (18 samples) and testing (80 samples) subsets, as in \cite{rony2019weak-loc-histo-survey}. It is important to note that the samples present a large variation in terms of gland shape, size, as well as overall H\&E stain (Fig.\ref{fig:glas-results-visu}).

\begin{figure}[h!]
  \center
\includegraphics[width=.6\linewidth]{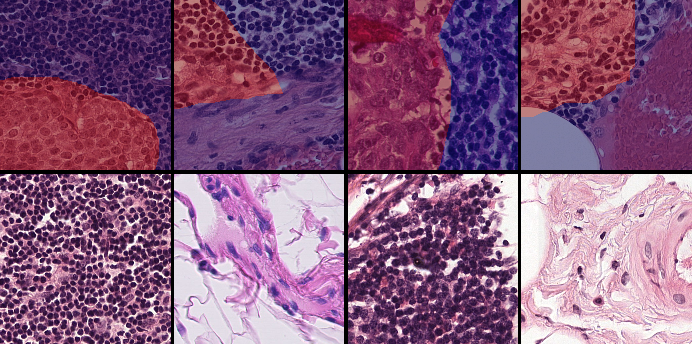}
\caption{Examples of testing samples from metastatic (top) and normal (bottom) classes of the Camelyon16-P512 dataset. Metastatic regions are indicated with a red mask. Best visualized with color.}
\label{fig:fig-cam16-samples}
\end{figure}

\noindent \textbf{(b) Camelyon16 patch-based benchmark \cite{rony2019weak-loc-histo-survey}:}
This benchmark is derived from the Camelyon16 dataset \cite{camelyon2016paper}, which contains 399 whole-slide images (normal or metastatic) for detection of metastases in H\&E stained tissue sections of sentinel auxiliary lymph nodes (SNLs) of women with breast cancer. In \cite{rony2019weak-loc-histo-survey}, the authors designed a protocol to sample patches with global and pixel-level annotations. Following this protocol, a patch can either be \textit{(i)} normal without any metastatic regions, \textit{(ii)} or metastatic with both normal and metastatic or only metastatic regions.
In this work, we consider the benchmark containing patches of size 512x512 (which we refer to as Camelyon16-P512), and use the same split as in \cite{rony2019weak-loc-histo-survey}. This benchmark contains a total of 48,870 samples: 24,348 samples for training, 8,858 samples for validation, and 15,664 samples for testing. Several examples from this data set are depicted in Fig. \ref{fig:fig-cam16-samples}).

\subsection{Evaluation Protocol}

\noindent \textbf{(a) Metrics:}
At the image level, we report the classification error,
\begin{equation}
    \label{eq:eq-cl-error}
    100 * (\frac{\#\text{misclassified samples}}{
    \#\text{samples}}) \; ,
\end{equation}
where ${\#\text{misclassified samples}}$ is the total number of miscalssified samples, and ${\#\text{samples}}$ is the total number of samples.

To report the pixel-level results, we resort to the F1 score, \ie Dice index, on the foreground region, referred to as F1$^+$,
\begin{equation}
    \label{eq:eq-f1+}
    F1^{+} = \frac{2  |\mathbb{G}^+ \cap \;\mathbb{S}^+|}{|\mathbb{G}^+| + |\mathbb{S}^+|} \; ,
\end{equation}
where $\mathbb{G}^+$ and $\mathbb{S}^+$ are the binary foreground masks of the ground-truth and prediction, respectively, and ${|\cdot|}$ the cardinality of a set. Furthermore, to measure how well the model is able to identify irrelevant regions, we also report the F1 score over the background, referred to as F1$^-$,
\begin{equation}
    \label{eq:eq-f1-}
    F1^{-} = \frac{2  |\mathbb{G}^- \cap \; \mathbb{S}^-|}{|\mathbb{G}^-| + |\mathbb{S}^-|} \; ,
\end{equation}
where $\mathbb{G}^-$ and $\mathbb{S}^-$ are the binary background masks of the ground-truth and prediction, respectively.

\noindent \textbf{(b) Reference methods:}
We compare our method to relevant state-of-the-art WSL segmentation approaches. These methods include:
CAM-Avg \cite{zhou2016learning}, CAM-Max \cite{oquab2015object}, CAM-LSE \cite{PinheiroC15cvpr,sun2016pronet}, WILDCAT \cite{durand2017wildcat}, Grad-CAM \cite{selvaraju2017grad}, Deep MIL \cite{ilse2018attention}, ERASE \cite{wei2017object} and the constrained convolutional network loss approach in \cite{kervadec2019constrained} (PN), where only the image-level label is used as supervision (i.e., \textit{presence} vs \textit{non-presence}). To establish an upper bound for the performance of pixel-level predictions, we train U-Net \cite{Ronneberger-unet-2015} under full supervision, i.e., all the pixel-level labels are known during training. Furthermore, to show that the F1$^+$ metric alone is not sufficient for evaluating pixel-level predictions, we employ a trivial mask prediction with all the pixels set to 1, as a lower baseline. We refer to our model in Eq. \eqref{eq:eq-5} as EEM when our max-uncertainty regularizer $\mathbf{R}$ is given by Eq. \eqref{eq:eq-6}, or as SEM when $\mathbf{R}$ is given by Eq. \eqref{eq:eq-7}.

\subsection{Implementation Details}
We employ a pre-trained ResNet18 CNN \cite{heZRS16} as backbone for feature extraction in all the methods, which is fine-tuned on the training set. ResNet is a common choice in the literature due to its high performance in image classification and, most importantly, time-efficiency. The choice of the specific model across the ResNet variants, i.e., ResNet18, is based on the number of trainable parameters. In particular, given the relatively small number of training samples available for the classification task (in comparison to natural images), lighter models are more robust against over-fitting than their higher-capacity counterparts. Most hyper-parameters are tuned empirically through validation or adopted from original works. All the methods are trained using SGD with the Nesterov momentum set to ${0.9}$, and a weight decay of ${10^{-4}}$ (and ${10^{-5}}$ for PN and ERASE methods).

All the weakly-supervised methods evaluated in this work, including ours, are trained using image-level annotations and, therefore, have no access to pixel-wise supervision during training. Only the fully supervised upper-bound (U-Net) uses pixel-wise labels during training.

\noindent \textbf{(a) Proposed approach:}
In our method, ${\mathcal{M}}$ and ${\mathcal{C}}$ share the same pre-trained backbone ResNet18 to avoid over-fitting and reduce model complexity.
\textbf{GlaS dataset}:
We use WILDCAT pooling for CAMs extraction \cite{durand2017wildcat} with its default hyper-parameters: $5$ class-related modalities per category, $kmax=0.3, kmin=0$, and dropout equal to $0.1$. Furthermore, ${\bm{M}^-}$ is set to be the complement of ${\bm{M}^+}$. ${\mathcal{M}}$ estimates ${\bm{M}^+}$ by selecting the winning CAM in WILDCAT pooling. To make this discrete selection differentiable, we employ a weighted sum of the CAMs with the class corresponding posterior probability ${\hat{p}}$ (Fig.\ref{fig:fig-proposal}). Additionally, we set empirically ${\lambda}=10^{-7}$ in \eqref{eq:eq-5}.  In \eqref{eq:eq-8}, we set $\sigma=0.15$ and $\omega=5$. Our method is trained for $80$ epochs with a batch size of $4$\footnote{Since our method uses two ResNet18 models, and it interpolates the mask into the full image size, it requires more GPU memory. Therefore, we used small batch size that can fit easily in standard GPUs with up to 11GB of memory.}. Regarding the log-barrier optimization in \eqref{eq:eq-5}, we set an initial ${t=5}$, which is increased by a multiplicative factor of ${1.01}$ after each epoch, up to a maximum value of ${10}$, following \cite{belharbi2019unimoconstraints,kervadec2019log}.
\textbf{Camelyon16-P512 benchmark}: The hyper-parameters used for this benchmark are set as follows. The number of training epochs is set to 20. The learning rate is set to $0.001$, and divided by 10 after performing half of the training epochs\footnote{The learning rate of the classification part is multiplied by 10 until reaching $0.01$.}. We set the batch size to 8. Patches are randomly rotated with an angle in $\{0^{\circ}, 90^{\circ}, 180^{\circ}, 270^{\circ}\}$. For WILDCAT, we set $kmax=kmin=0.1$, class-related modalities to $4$, and $\alpha=0.6$ \cite{rony2019weak-loc-histo-survey}. 
For the log-barrier method, we initialized $t=10$, and increased it by a multiplicative factor of $1.1$ per epoch, until a maximum value of $30$. The rest of the hyper-parameters are kept as in the GlaS dataset. We provide in the appendix an additional ablation study to motivate the choice of the hyper-parameters employed in our method.

\noindent \textbf{(b) Prior WSL approaches:}
\textbf{GlaS dataset}:
The pooling setup employed in PN and ERASE approaches to extract the CAMs is similar to the one we used in our method, i.e., WILDCAT. We set the $\lambda$ value defined in PN \cite{kervadec2019constrained} to $10^{-5}$. Regarding the ERASE approach, we follow \cite{wei2017object} and erase the pixels that belong to the top ${20\%}$ of the maximum activation. Both approaches are trained using a batch size of $8$, for $400$ epochs, with a learning rate of ${0.001}$, which is decayed every $40$ epochs by $0.1$, up to a minimum learning rate of $10^{-7}$. The remaining WSL techniques are trained using the hyper-parameters in \cite{rony2019weak-loc-histo-survey}. More specifically, the training is conducted for $160$ epochs, with a learning rate of ${0.01}$, which is reduced to ${0.001}$ in the last ${80}$ epochs. Furthermore, the batch size is set to $32$. In all the WSL methods, we used the validation classification error as a stopping criterion.
\textbf{Camelyon16-P512 benchmark}: The number of epochs, learning rate, batch size and data-augmentation strategy are the same as in our approach. The remaining hyper-parameters are kept the same as in the GlaS data set.

\noindent \textbf{(c) Full supervision:}
The upper bound, i.e., fully-supervised segmentation with pixel-level labels, is achieved using
U-Net \cite{Ronneberger-unet-2015} on both datasets.
\textbf{GlaS dataset}:
Network training is performed with SGD for $960$ epochs, with the Nesterov momentum parameter set to $0.9$. We used a value of $10^{-4}$ for
weight decay, a value of $16$ for batch size and a value of $0.1$ for the initial learning rate, with the latter divided by $10$ every $320$ epochs. \textbf{Camelyon16-P512 benchmark}: The training is conducted for 90 epochs, with a learning rate set initially to $0.1$ and
divided by 10 every 30 epochs. The rest of the hyper-parameters are identical to those used for the GlaS data set.

In all the methods, input patches were resized to ${416\times416}$ prior to data augmentation based on random rotation and horizontal/vertical flipping. To increase robustness of all models to stain variations, we used random jittering\footnote{Jittering transform is part of the package torchvision of Pytorch library \url{https://pytorch.org}.} (brightness=$0.5$, contrast=$0.5$, saturation=$0.5$ and hue=$0.05$) over the images.

\subsection{Results}

\begin{figure*}[h!]
  \centering
  \includegraphics[width=1.\linewidth]{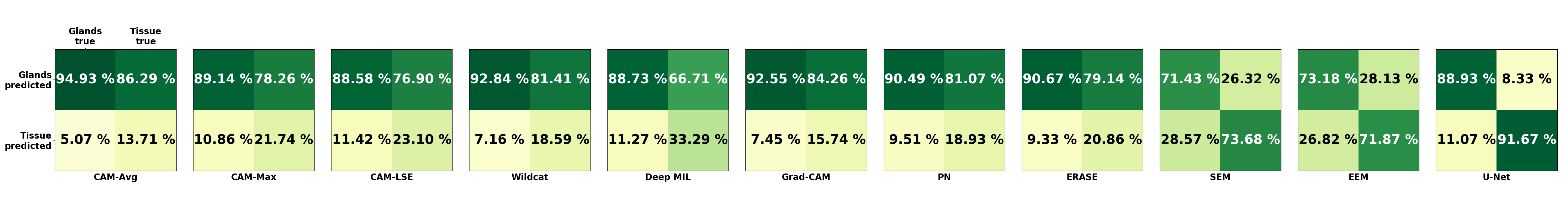}
  \caption{\textbf{GlaS dataset}: Confusion matrix over entire pixels of test set. (Best visualized in color.)}
  \label{fig:glas-cmatrix}
\end{figure*}

\noindent \textbf{(a) Quantitative results:}
Regarding the performance in terms of segmentation, results on the GlaS dataset are reported in Tab. \ref{tab:tab3-glas} while Camelyon16-P512 results are presented in Tab. \ref{tab:tab3-camelyon16-p512}. On the GlaS dataset, we can observe that our method achieves the best F1$^+$ score among all the WSL approaches, providing a boost in performance of nearly $3.5\%$ with respect to the second best performing method, i.e., Deep MIL. An important observation is that predicting an entire mask with all the pixels set to 1 yields an F1$^+$ score of $66.01\%$, which motivates the use of the additional F1${^-}$ metric. While all the WSL approaches obtain F1${^-}$ values below $30\%$, except Deep MIL ($41.34\%$), our method achieves an F1${^-}$ of ${\sim 69\%}$,
suggesting that the proposed KL regularization helps substantially in localizing the background. It brings a significant decrease of false positives for pixel-wise predictions, yielding more reliable segmentations. Fig. \ref{fig:glas-cmatrix} depicts the confusion matrices over all the pixels of the testing set.
In particular, we observe that, although the existing methods can generally identify the gland pixels, they often fail to correctly identify normal-tissue pixels. As we will see in the qualitative results, this yields over-segmentations. Unlike the existing methods, our approach correctly identified substantially larger amounts of tissue pixels. We note that explicitly maximizing the entropy (EEM) and resorting to the KL surrogate (SEM) yield very similar performances, despite the different gradient dynamics.

\vspace{0.25cm}
Note that the ERASE algorithm in \cite{wei2017object} typically yielded masks with all the pixels set to 1, which prevented from using the online prohibitive segmentation learning (PSL) algorithm \cite{wei2017object} to discover more complete regions. The size-constraint approach in \cite{kervadec2019constrained} achieved similar low performance because the use of the \textit{presence} and \textit{non-presence} constraints does not impose any upper bound on the size of the regions of interest, which, typically, results in the activation of large regions.

\vspace{0.25cm}

\begin{figure*}[h!]
  \centering
  \includegraphics[width=1.\linewidth]{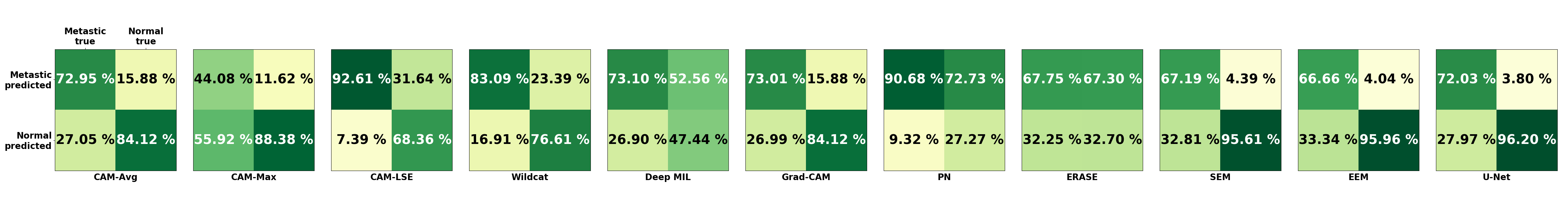}
  \caption{\textbf{Camelyon16-P512 dataset}: Confusion matrix over entire pixels of test set. (Best visualized in color.)}
  \label{fig:camelyon16-cmatrix}
\end{figure*}

On the Camelyon16-P512 dataset, most recent WSL method yield $F1^{+}$ scores higher than 60$\%$, but the $F1^{-}$ values are below 80$\%$,
 except CAM-Max. The latter tends to predict very small tumorous regions, leading to very low $F1^{+}$ and high $F1^{-}$ scores. By inspecting the pixel-wise confusion matrices in Fig.\ref{fig:camelyon16-cmatrix}, we notice that, despite improving their performances in comparisons to those obtained for the GlaS dataset, most of the existing WSL methods struggle to provide a good balance between true positive/negatives. When tumors are properly segmented, normal tissues are poorly identified, and vice-versa.
On the contrary, our method obtained a good balance, showing a behaviour close to the fully-supervised model.

\vspace{0.25cm}

The performances in terms of image classification are provided in the image-level columns in Tables \ref{tab:tab3-glas} and \ref{tab:tab3-camelyon16-p512}. The proposed method obtained the lowest classification error on the GlaS dataset, similarly to CAM-avg\cite{zhou2016learning} and Grad-CAM \cite{selvaraju2017grad}. On the Camelyon16-P512 benchmark, Wildcat yielded the best error, i.e., $1.48\%$, and our model achieved a lower performance, but is still competitive in comparison to other methods. It is worth
mentioning that the classification task in our method is in direct competition with other additional constraints, e.g., Eq. \ref{eq:eq-5}, which could drive optimization to favor a compromise solution over all the sub-tasks. Note that, while fully-supervised
U-Net provides the best segmentation results on both datasets (due to access to pixel-level labels), it cannot provide image-class predictions simultaneously.

\vspace{0.25cm}

\begin{figure}[h!]
  \centering
  \includegraphics[width=1.\linewidth]{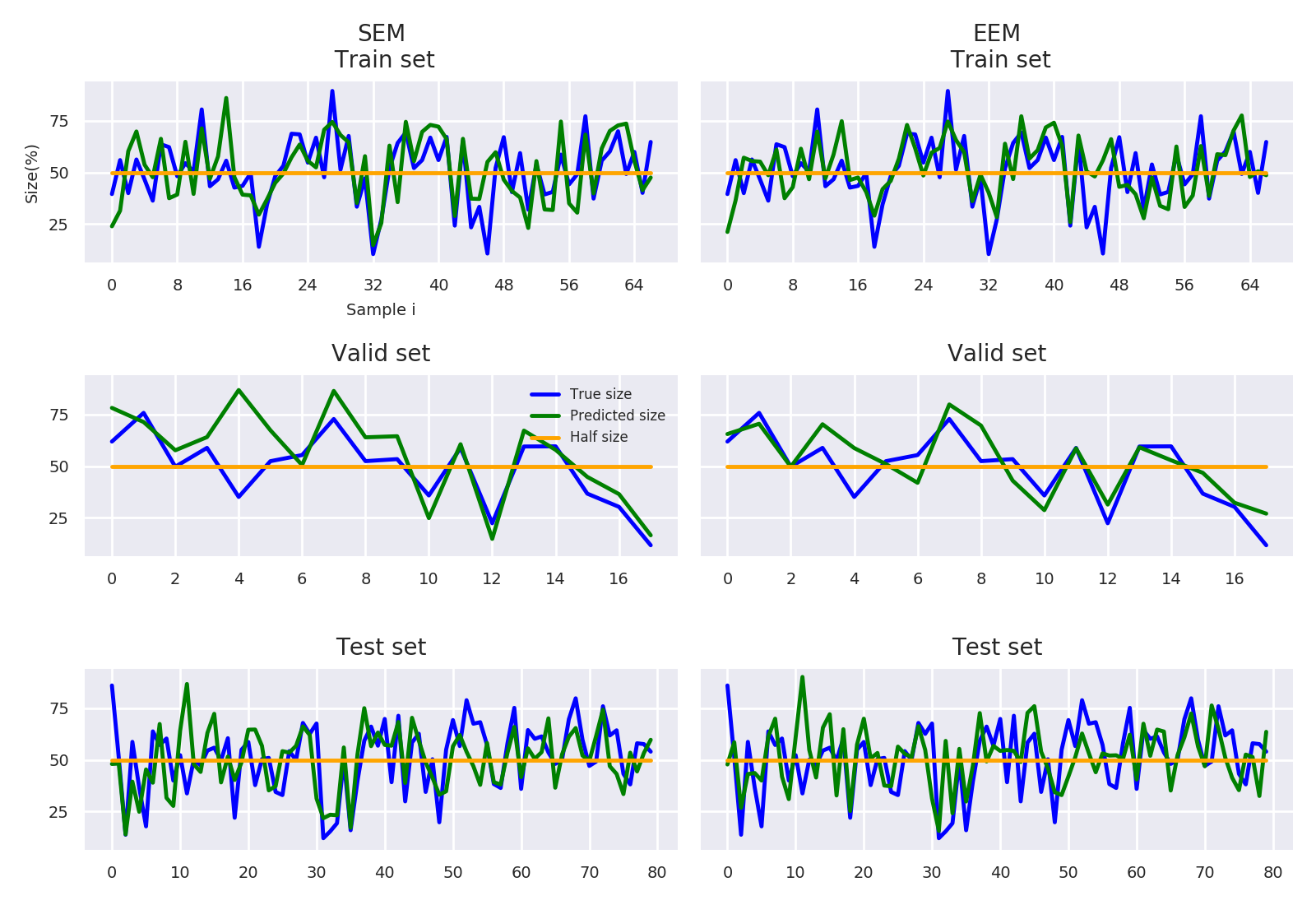}
  \caption{\textbf{GlaS dataset}: Comparison between the foreground-mask sizes: region size corresponding to the true mask (blue), predicted mask (green) and perfectly balanced solution, i.e., half of the image size (orange) over the training/validation/testing sets of the GlaS dataset, for SEM and EEM methods. The region size in a sample is normalized by the total number of pixels. The x-axis is the sample identifier. The y-axis is region size expressed in terms of percentage with respect to the total
  image area. (Best visualized in color.)}
  \label{fig:glas-ablation-size}
\end{figure}

\noindent \textbf{(b) Size constraints:}
To examine experimentally the effect of the log-barrier balancing term in Eq. \ref{eq:eq-5}, we plotted in Fig.\ref{fig:glas-ablation-size} the predicted foreground-region sizes along with the ground-truth sizes, and juxtapose these to the perfectly balanced solution (i.e., half the size of the image), over the training, validation and testing sets. These results suggest that using the log-barrier soft penalties enables to predict masks that are consistent with the true region sizes, and do not necessarily correspond to perfectly balanced solutions, while discouraging trivial, extremely imbalanced segmentation. The figure also shows the large variation in the sizes of the target ground-truth regions, which makes perfectly balanced solutions less  likely  to  be  optimal  for  all  the training samples.

\vspace{0.25cm}

\begin{figure}[h!]
  \centering
  \includegraphics[width=1.\linewidth]{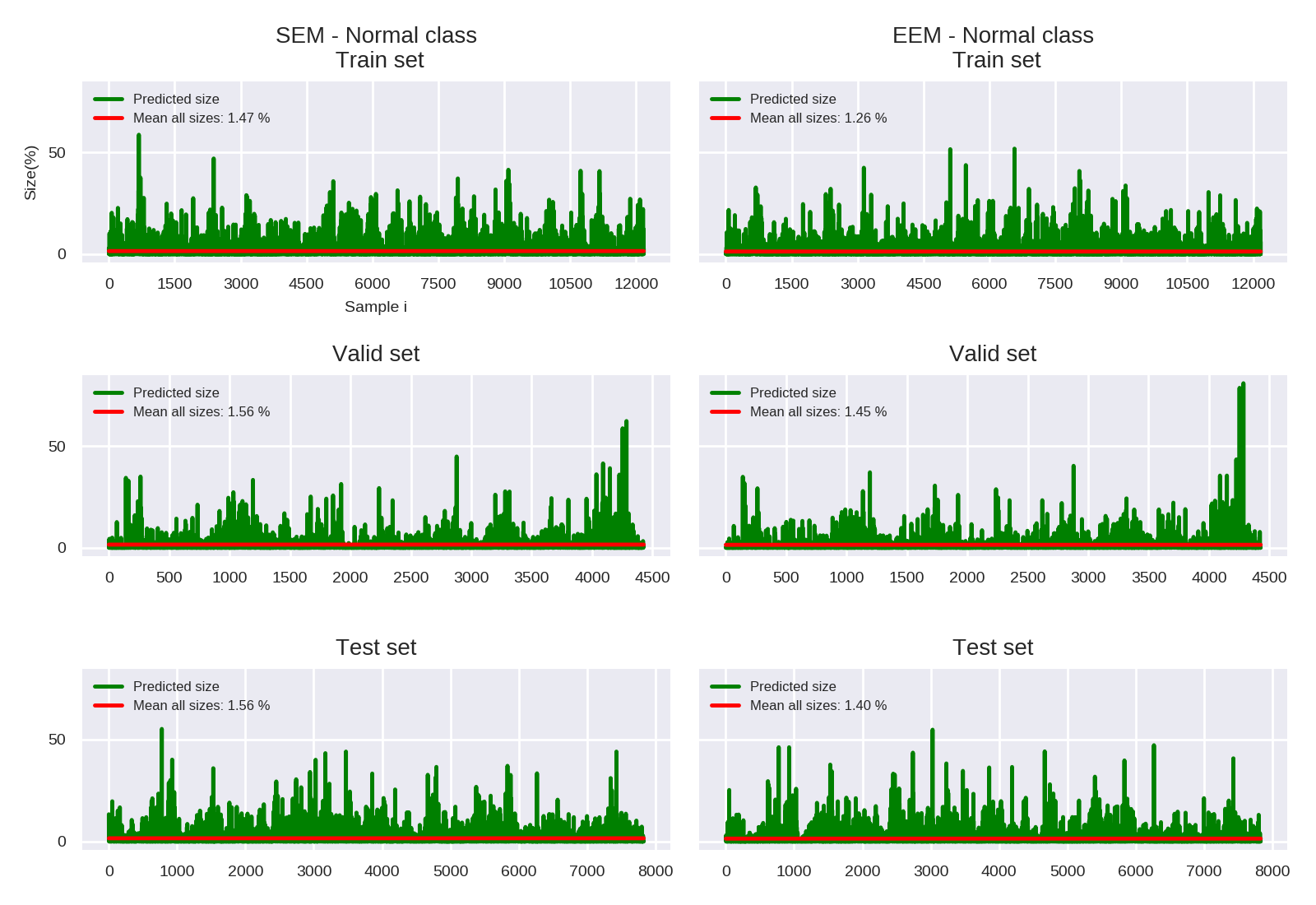}
  \caption{\textbf{Camelyon16-P512 benchmark}: Region size corresponding to the predicted mask over the \textbf{normal} samples (green) of the  training/validation/testing sets of the Camelyon16-P512 benchmark, for SEM/EEM methods. Note that normal samples do not contain any metastatic regions (i.e. the size of the foreground must be zero). The average size over all subsets is less than $1.56\%$.
  Size in a sample is normalized by the total number of pixels.
  The x-axis is the sample identifier. The y-axis is region size expressed in terms of percentage with respect to the total image area. (Best visualized in color.)}
  \label{fig:cam16-normal-ablation-size}
\end{figure}

Over the Camelyon16-P512 dataset, we illustrate with Fig.\ref{fig:cam16-normal-ablation-size} an interesting benefit of the log-barrier loss over normal samples.
In these samples, there is no tumorous regions, \ie, no foreground. The log-barrier enables to increase the size of the background region, thereby suppressing most of the false positives. The average size of tumorous regions in normal samples is ${< 1.56\%}$. This shows that our unsupervised log-barrier size penalty provides a helpful tool to deal
with false positives (large foreground regions) in these samples.

\vspace{0.50cm}

\begin{table}[h!]
  \caption{Image-level classification and pixel-level segmentation performances on the GlaS test set. Cl: classification. The best performance is shown in bold.
  }
  \label{tab:tab3-glas}
  \centering
  \small
   \resizebox{1.\linewidth}{!}{
  \begin{tabular}{l|r|r|r}
    \toprule
    &  \multicolumn{1}{c|}{\textbf{Image level}} & \multicolumn{2}{c}{\textbf{Pixel level}} \\
    \textbf{Method} &  \multicolumn{1}{c|}{Cl. error (\%)} & \multicolumn{1}{c|}{F1$^+$ (\%)} &  \multicolumn{1}{c}{F1$^-$ (\%)}\\
    \midrule
    All-ones  (Lower-bound)                                                & $--$        & $66.01$      & $00.00$   \\
    \midrule
    PN \cite{kervadec2019constrained}                       & $--$        & $65.52$      & $24.08$  \\
    ERASE \cite{wei2017object}                              & $7.50$       & $65.60$      & $25.01$  \\
    CAM-Max  \cite{oquab2015object}                         & $1.25$      & $66.00$      & $26.32$ \\
    CAM-LSE  \cite{PinheiroC15cvpr,sun2016pronet}           & $1.25$      & $66.05$      & $27.93$  \\
    Grad-CAM \cite{selvaraju2017grad}                       & $\bm{0.00}$ & $66.30$      & $21.30$  \\
    CAM-Avg \cite{zhou2016learning}                         & $\bm{0.00}$ & $66.90$      & $17.88$  \\
    Wildcat \cite{durand2017wildcat}                        & $1.25$      & $67.21$      & $22.96$  \\
    Deep MIL \cite{ilse2018attention}                       & $\bm{2.50}$ & $68.52$      & $41.34$  \\
    \midrule
    Ours (EEM)                                   & $\bm{0.00}$ & $\bm{72.11}$ & $69.07$   \\
    Ours (SEM)                                   & $\bm{0.00}$ & $71.94$ & $\bm{69.23}$   \\
    \midrule
    U-Net \cite{Ronneberger-unet-2015}   (Upper-bound)        & $--$        & $90.19$      & $88.52$ \\
    \bottomrule
  \end{tabular}
}
\end{table}

\vspace{2.5cm}

\begin{table}[h!]
  \caption{Image-level classification and pixel-level segmentation performances on the Camelyon16-P512 test set. Cl: classification. The best performance is shown in bold.
  }
  \label{tab:tab3-camelyon16-p512}
  \centering
  \small
   \resizebox{1.\linewidth}{!}{
  \begin{tabular}{l|r|r|r}
    \toprule
    &  \multicolumn{1}{c|}{\textbf{Image level}} & \multicolumn{2}{c}{\textbf{Pixel level}} \\
    \textbf{Method} &  \multicolumn{1}{c|}{Cl. error (\%)} & \multicolumn{1}{c|}{F1$^+$ (\%)} &  \multicolumn{1}{c}{F1$^-$ (\%)}\\
    \midrule
    All-ones  (Lower-bound)                                 & $--$        & $59.44$      & $00.00$   \\
    \midrule
    PN \cite{kervadec2019constrained}                       & $--$        & $31.15$      & $37.36$  \\
    ERASE \cite{wei2017object}                              & $8.61$      & $31.30$      & $42.48$  \\
    CAM-Max  \cite{oquab2015object}                         & $10.06$     & $48.28$      & $81.92$  \\
    CAM-LSE  \cite{PinheiroC15cvpr,sun2016pronet}           & $1.51$      & $64.31$      & $63.78$  \\
    Grad-CAM \cite{selvaraju2017grad}                       & $2.40$      & $62.78$      & $79.05$  \\
    CAM-Avg \cite{zhou2016learning}                         & $2.40$      & $62.75$      & $79.05$  \\
    Wildcat \cite{durand2017wildcat}                        & $\bm{1.48}$ & $62.73$      & $72.59$  \\
    Deep MIL \cite{ilse2018attention}                       & $1.93$      & $59.01$      & $36.94$  \\
    \midrule
    Ours (EEM)                                   & $6.26$ & $67.98$ & $\bm{88.80}$   \\
    Ours (SEM)                                   & $6.95$ & $\bm{68.26}$ & $88.55$   \\
    \midrule
    U-Net \cite{Ronneberger-unet-2015}   (Upper-bound)        & $--$        & $71.11$      & $89.68$ \\
    \bottomrule
  \end{tabular}
}
\end{table}

\begin{table}[h!]
  \caption{Complexity of different methods in terms of: a) total number of learnable parameters (\#pModel) and b) average inference time per image (aInfTime) considering only the forward phase. The used backbone is ResNet18. Computations are performed on an idle GPU GeForce GTX-1080-Ti on the GlaS test set. Image size is ${522 \times 775}$.}
  \label{tab:tab3-glas-complexity}
  \centering
  \small
   \resizebox{1.\linewidth}{!}{
  \begin{tabular}{l|r|r}
    \toprule
    \textbf{Method} &  \multicolumn{1}{c|}{\#pModel} & \multicolumn{1}{c|}{aInfTime (sec/image)}\\
    \midrule
    PN \cite{kervadec2019constrained}                       & $11,330,122$       & $00.0034$   \\
    ERASE \cite{wei2017object}                              & $11,326,016$       & $00.0027$  \\
    CAM-Max  \cite{oquab2015object}                         & $11,177,538$       & $00.3111$  \\
    CAM-LSE  \cite{PinheiroC15cvpr,sun2016pronet}           & $11,177,538$       & $00.3790$   \\
    Grad-CAM \cite{selvaraju2017grad}                       & $11,177,538$       & $05.2132$   \\
    CAM-Avg \cite{zhou2016learning}                         & $11,177,538$       & $00.3295$   \\
    Wildcat \cite{durand2017wildcat}                        & $11,180,616$       & $00.3596$   \\
    Deep MIL \cite{ilse2018attention}                       & $11,243,460$       & $00.4523$   \\
    \midrule
    Ours (EEM, SEM)                                         & $11,335,252$       & $00.0190$   \\
    \midrule
    U-Net \cite{Ronneberger-unet-2015}                      & $14,154,706$       & $00.0081$ \\
    \bottomrule
  \end{tabular}
}
\end{table}

\begin{table}[h!]
  \caption{Ablation study for our model over the GlaS test set.
  FG: foreground \eqref{eq:eq-1-1}.
  BG: Background.
  EEM: Eq. \eqref{eq:eq-6}.
  SEM: Eq. \eqref{eq:eq-7}.
  ASC: The absolute size constraints in Eq. \eqref{eq:eq-5}.
  }
  \label{tab:tab-ablation-study}
  \centering
  \small
  \resizebox{0.95\linewidth}{!}{
  \begin{tabular}{l|r|r|r}
    \toprule
    &  \multicolumn{1}{c|}{\textbf{Image level}} & \multicolumn{2}{c}{\textbf{Pixel level}} \\
    Method &  \multicolumn{1}{c|}{Cl. error (\%)} & \multicolumn{1}{c|}{F1$^+$ (\%)} &  \multicolumn{1}{c}{F1$^-$ (\%)}\\
    \midrule
    Wildcat \cite{durand2017wildcat}    & $1.25$      & $67.21$      & $22.96$  \\
    \midrule
    FG only                 & $1.25$      & $71.54$      & $49.23$  \\
    \midrule
    FG + BG (EEM)           & $1.25$      & $72.54$      & $61.82$  \\
    FG + BG (EEM) + ASC     & $0.00$      & $72.11$      & $69.07$  \\
    \midrule
    FG + BG (SEM)           & $1.25$      & $\bm{72.96}$      & $61.95$  \\
    FG + BG (SEM) + ASC     & $0.00$      & $71.94$      & $\bm{69.23}$  \\
    \bottomrule
  \end{tabular}
  }
\end{table}

\begin{figure}[h!]
  \centering
  \includegraphics[width=.8\linewidth]{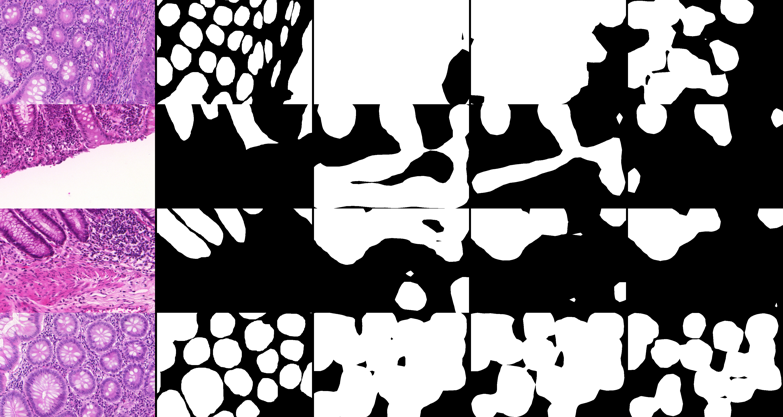} \\
  \vspace{0.2cm}
  \includegraphics[width=.8\linewidth]{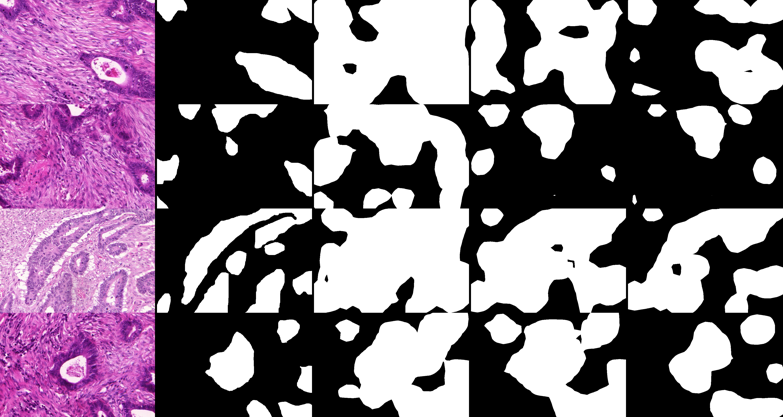} \\
  \includegraphics[width=.8\linewidth]{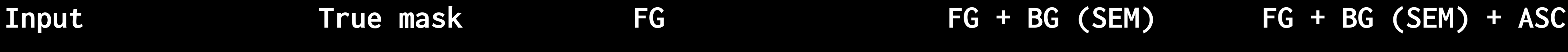}
  \caption{Visual results of the ablation study over GlaS test set. Top samples: Benign. Bottom samples: Malignant.
  From left to right: Input image, True mask, FG, FG + BG (SEM), FG + BG (SEM) + ASC. (Best visualized in color.)}
  \label{fig:glas-results-visu-ablation}
\end{figure}

\begin{figure}[h!]
  \centering
  \includegraphics[width=1.\linewidth]{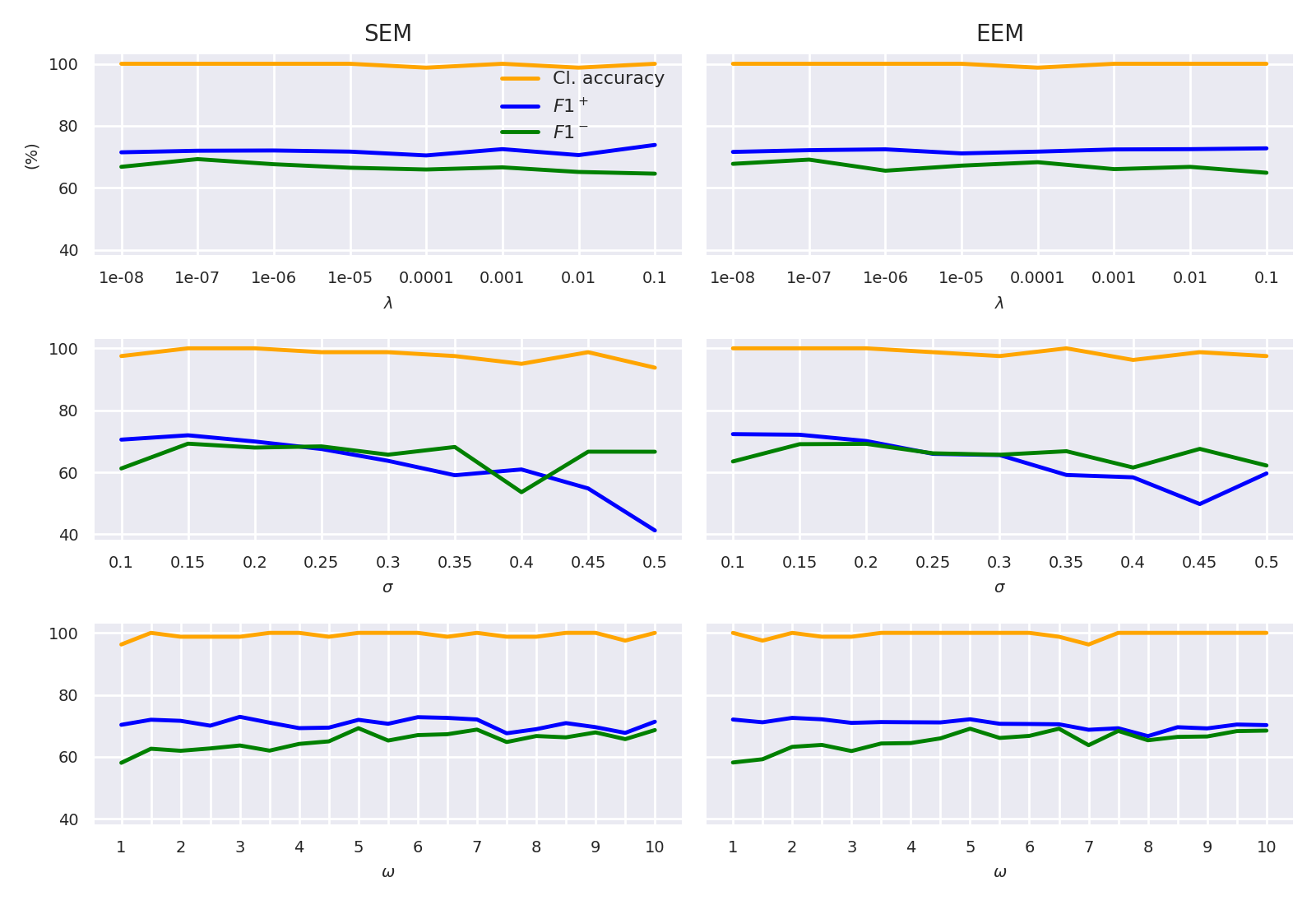}
  \caption{Ablation study with respect to the main three different hyper-parameters of our method (from top to bottom: ${\lambda, \sigma, \omega}$) on the GlaS testing set. \textbf{Left}: SEM method. \textbf{Right}: EEM method. The \textit{classification accuracy} is in orange, whereas segmentation metrics are indicated in green and blue. (Best visualized in color.)}
  \label{fig:glas-ablation-hyper-param}
\end{figure}

\begin{figure*}[h!]
  \centering
  \includegraphics[width=.9\linewidth]{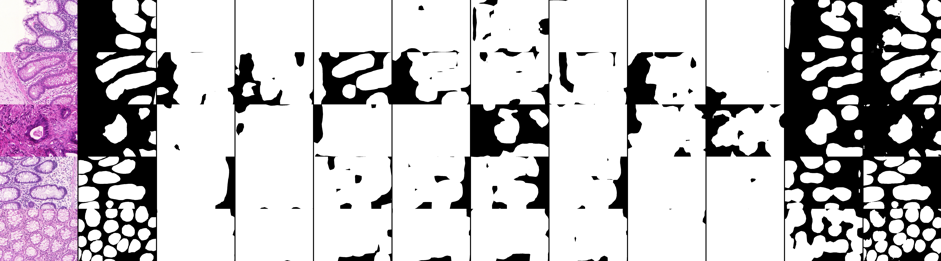} \\
  \includegraphics[width=.9\linewidth]{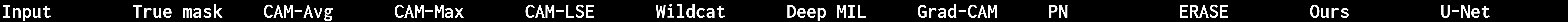}
  \caption{\textbf{GlaS dataset}: Qualitative results of the predicted binary mask for each method on several GlaS test images. Our method, referred to as \textit{Ours}, is the SEM version with the ASC regularization term. (Best visualized in color.)}
  \label{fig:glas-results-visu}
\end{figure*}

\begin{figure*}[h!]
  \centering
  \includegraphics[width=.9\linewidth]{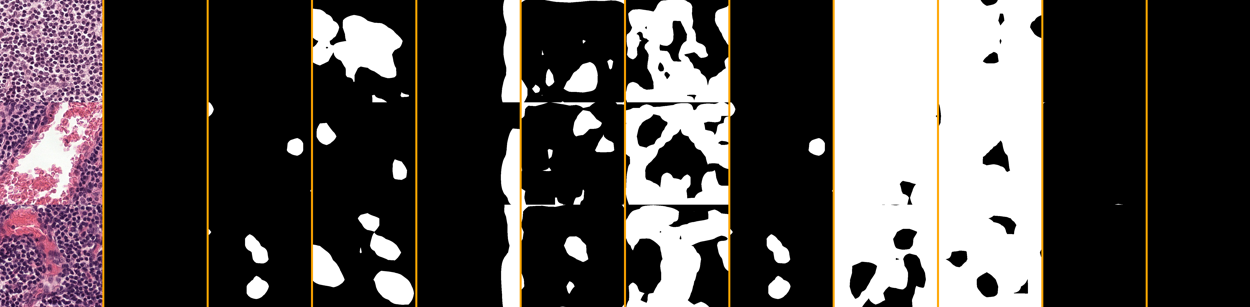} \\
  \includegraphics[width=.9\linewidth]{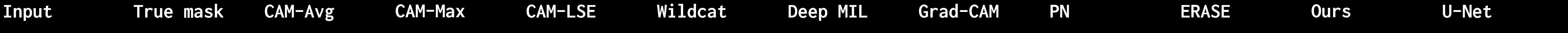}
  \caption{\textbf{Camelyon16-P512 benchmark}: Examples of mask predictions over \textbf{normal} samples from the testing set. White pixels indicate metastatic regions, while black pixels indicate normal tissue. This illustrates false positives. Note that normal samples do not contain any metastatic regions. Ours is SEM version with the ASC regularization. (Best visualized in color.)}
  \label{fig:cam16-results-visu-normal-patch}
\end{figure*}

\begin{figure*}[h!]
  \centering
  \includegraphics[width=.9\linewidth]{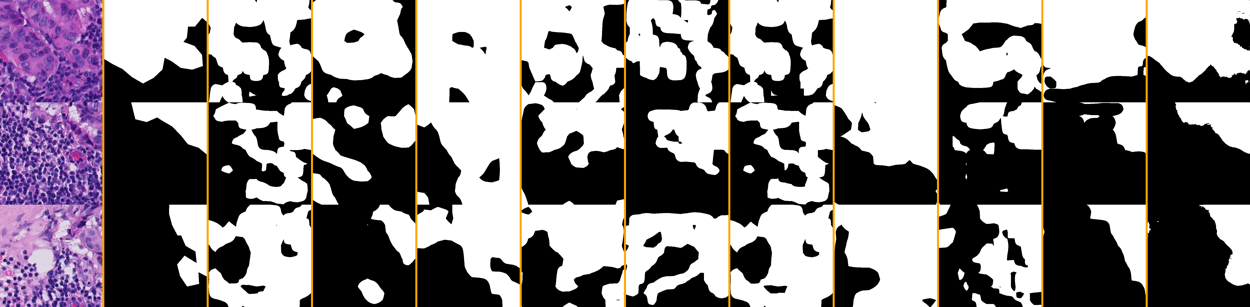} \\
  \includegraphics[width=.9\linewidth]{code-camelyon16-512-patch}
  \caption{\textbf{Camelyon16-P512 benchmark}: Examples of predicted pixel-wise masks over \textbf{metastatic} samples from the test set. White pixels indicate metastatic regions, while black pixels represent normal tissue. \textit{Ours} is the SEM version with the ASC regularization. (Best visualized in color.)}
  \label{fig:cam16-results-visu-metastatic-patch}
\end{figure*}

\noindent \textbf{(c) Complexity}:
Tab. \ref{tab:tab3-glas-complexity} reports the complexity of each method in terms of the number of parameters and inference time per image.
Based on ResNet18 as backbone architecture, all the weakly-supervised methods have the same number of parameters (${\sim 11}$ millions parameters). The slight difference between the models stems from the output pooling layer that may include additional parameters depending on the method. Also, all the methods perform the inference step in less than a second on an average GPU. It is worth noting that, under the same conditions, our inference is substantially faster (typically one order of magnitude) than the weakly-supervised methods under evaluation.

\noindent \textbf{(d) Ablation studies:}

\emph{Method components}: To evaluate the impact of our regularization terms, we conducted additional experiments over GlaS. As we employ the pooling operations proposed in WILDCAT \cite{durand2017wildcat}, the latter is the closest approach to our method without the KL regularization term. Nevertheless, there is a main difference: WILDCAT \cite{durand2017wildcat} implicitly predicts the foreground mask ${\bm{M}^+}$ through classification, whereas our model explicitly predicts the foreground mask and performs classification only over the regions identified by the mask.
As reported in Tab.\ref{tab:tab-ablation-study}, explicitly modeling the foreground results in better segmentations, which leads to a further decrease of false positives. Particularly, while the F1$^+$ score increases by nearly 4\%, the improvement is about 27\% in terms of F1$^-$. A similar trend is observed when we incorporate the proposed regularization terms, i.e., EEM and SEM, to describe the background. In this case, the improvement on the foreground is marginal, with values ranging from 1.0\% to 1.5\%, whereas the background segmentation improvement is enhanced by nearly 12\%. This can be explained by the fact that this term pushes explicitly the model to be aware of the presence of the background.

We also investigate the impact of integrating the size-balancing constraints in our formulation, which is denoted as Absolute Size Constraints (ASC). We observe that seeking the largest regions for both the background and foreground using an absolute size constraint improves the background segmentation measure (F1$^-$), while maintaining the foreground segmentation performance (F1$^+$). It promotes region balancing due to the competition between the foreground and background via the size constraints, which prevent the dominance of one region over the other. More specifically, adding the log-barrier size constraints in (\ref{eq:eq-5}) brought a boost of nearly 8\% in the F1$^-$ metric, compared to the model without the ASC term. Notice that, in addition to improving the background segmentation performance, the ASC term helps reducing the image classification error.

We also depict the visual results in Fig.\ref{fig:glas-results-visu-ablation}. First, one can observe that including the proposed background regularization term, i.e., FG + BG (SEM), reduces over-segmentations of the target regions, in comparison to using only the foreground term. Second, integrating the absolute size constraints (\textit{last column}) into the learning objective reduces the number of false-positive predictions. These visual results are in line with the quantitative results reported previously.

\emph{Hyper-parameters}:
We conducted additional experiments to assess the sensitivity of our method to the variations of the three main hyper-parameters: $\lambda$, $\sigma$ and $\omega$. In our experiments, these hyper-parameters
are varied uniformly within the following intervals:
$\lambda \in [1e^{-7}, 1e^{-1}]$, $\sigma \in [0.1, 0.5]$ and
$\omega \in [1, 10]$.  Fig.\ref{fig:glas-ablation-hyper-param} depicts the performances versus the hyper-parameters.

\vspace{0.5cm}
In terms of segmentation, the results suggest that our method is less sensitive to $\lambda$, whereas smaller variations in $\sigma$ and $\omega$ may result in performance changes. In particular, in the case of $\sigma$, our method achieves better $F1^+$ and $F1^-$ scores with small values. Note that $\sigma$ is a threshold that controls the flow of information to the foreground and background masks. Thus, high threshold values may prevent learning, particularly at the start of training where the activations are very low. Small threshold values are more preferable, enabling better information flow through all training epochs. The method is less sensitive to $\omega$, in comparison to $\sigma$. We recall that this is a heating hyper-parameter, typically set to values greater than 1, so as to amplify the activations. In term of classification, the influence of the hyper-parameters on performance is negligible.

\noindent \textbf{(e) Qualitative results:}
Fig. \ref{fig:glas-results-visu}, \ref{fig:cam16-results-visu-normal-patch}, \ref{fig:cam16-results-visu-metastatic-patch} depict representative visual results of the different approaches. We observe that all the prior WSL methods tend to yield over-segmentations, highlighting almost the entire image as a discriminative part. This could be observed in particular on the GlaS dataset (Fig \ref{fig:glas-results-visu}), and on Camelyon16-P512 over normal patches (Fig.\ref{fig:cam16-results-visu-normal-patch}) where discriminative regions easily overflow to normal tissue. In contrast, our approach identifies relevant regions more accurately. For instance, on the GlaS dataset, pathologists rely on the glands to assess colon cancer, whereas the rest of the image is not relevant for diagnosis. The visual results suggest that our approach enables to classify an image into benign/malignant based solely on gland information, similarly to pathologists. Additionally, an interesting finding is that our approach can handle multiple instances successfully, which is very convenient when several glands are present in the image. For this critical task, this makes our method more reliable 
than prior WSL techniques. Over Camelyon16-P512, and particularly over normal patches, our method tends to indicate that the entire image is tumor free. Nevertheless, prior WSL methods still predict tumor regions, although the image does not present any evidence supporting that.

\section{Conclusion}
\label{sec:conclusion}
Standard WSL methods based on global annotations may yield high false-positive rates when dealing with challenging images, with high visual similarities between the foreground and background regions. This is the case of histology images \cite{rony2019weak-loc-histo-survey}. The vulnerability to false positives comes mainly from the use of a discriminative training loss, which enables the emergence of regions of interest without pixel-wise supervision. In this work, we attempted to alleviate this vulnerability by explicitly constraining the model to be aware of the presence of the background (i.e., non-discriminative regions), which, to our knowledge, has not been considered before in WSL methods. We propose a principled uncertainty definition of the background, which is the largest part of the image where the model is most uncertain about its prediction of the image label. This definition is formulated as penalty terms to be optimized during training, with KL regularizers modeling high uncertainty. The obtained empirical results over GlaS and Camelyon16-P512 benchmarks showed the benefits of our proposal in term of segmentation accuracy and low false-positive rates, while maintaining competitive classification performances in comparison to state-of-the-art WSL techniques. Moreover, our method could be readily used for other medical applications. As a future direction, we consider extending our approach to handle multiple classes within the image. Different constraints could be applied over the predicted masks, such as texture properties, shape or other regional constraints.

\section*{Acknowledgment}
This research was supported in part by the Canadian Institutes of Health Research, the Natural Sciences and Engineering Research Council of Canada, and Compute Canada.

%
%
%
%

\FloatBarrier

\renewcommand{\theequation}{\thesection.\arabic{equation}}
\setcounter{equation}{0}

\renewcommand\thefigure{\thesection.\arabic{figure}}
\setcounter{figure}{0}

\appendices

\section{The different variants of the Kullback–Leibler divergence and their gradients for binary classification}
\label{sec:sec-app-0}

We show here the difference between ${\kl{\hat{p}^-}{q}}$ and ${\kl{q}{\hat{p}^-}}$, with $q$ being the uniform distribution, i.e., all the components of probability simplex vector $q$ are equal to $1/c$. The difference between these two max-uncertainty losses is due to the asymmetry of the Kullback–Leibler divergence.

\begin{align}
\kl{\hat{p}^-}{q} & = \sum_{l=1}^{c} \hat{p}_l^- \log\left(\frac{\hat{p}^-_l}{q_l}\right) \nonumber \\
 & = \sum_{l=1}^{c} \hat{p}_l^- \log\hat{p}_l^- - \sum_{l=1}^{c} \hat{p}_l^- \log \frac{1}{c}  \nonumber \\
 & = \sum_{l=1}^{c} \hat{p}_l^- \log\hat{p}_l^- + \log(c) \underbrace{\sum_{l=1}^{c} \hat{p}_l^-}_{=1}   \nonumber \\
 & = -\mathbf{H}(\hat{p}^-) + \log(c) \nonumber \\
 & \ceq -\mathbf{H}(\hat{p}^-) \label{eq2app}
\end{align}
where subscript $l$ in $p_l$ denotes the $l^{\mbox{{\tiny th}}}$-component of probability simplex vector $p$.

\begin{align}
\kl{q}{\hat{p}^-} & = \sum_{l=1}^{c} q_l \log\left(\frac{q_l}{\hat{p}_l^-}\right) \nonumber \\
 & = \underbrace{\sum_{l=1}^{c} q_l \log q_l}_{- \log(c)} - \underbrace{\sum_{l=1}^{c} q_l \log \hat{p}_l^-}_{\frac{1}{c}\sum_{l=1}^{c} \log \hat{p}_l^-}   \nonumber \\
 & \ceq \mathbf{H}(q, \hat{p}^-)
\end{align}

\textbf{Derivatives for binary classification}:

\begin{align}
\frac{\partial-\mathbf{H}(\hat{p}^-)}{\partial\hat{p}_1^-} & = \frac{\partial \left(\hat{p}_1^- \log \hat{p}_1^- + (1- \hat{p}_1^-)\log (1 - \hat{p}_1^-) \right)}{\partial \hat{p}_1^-} \label{eq9app} \\
 & = \log \hat{p}_1^-  - \log (1 - \hat{p}_1^-) \label{eq10app} \\
 & = \log\left(\frac{\hat{p}_1^-}{1 - \hat{p}_1^-}\right). \label{eq11app}
\end{align}

\begin{align}
\frac{\partial-\mathbf{H}(q, \hat{p}^-)}{\partial\hat{p}^-} & = \frac{\partial \left(-\frac{1}{2} (\log \hat{p}_1^- + \log (1 - \hat{p}_1^-)) \right)}{\partial \hat{p}_1^-} \label{eq12app} \\
 & = -\frac{1}{2} \left(\frac{1}{\hat{p}_1^-}  - \frac{1}{1 - \hat{p}_1^-}\right). \label{eq13app}
\end{align}

\section{EEM/SEM Training curves}
\label{sec:training-curves}
\begin{figure}[h!]
  \center
\includegraphics[width=1.\linewidth]{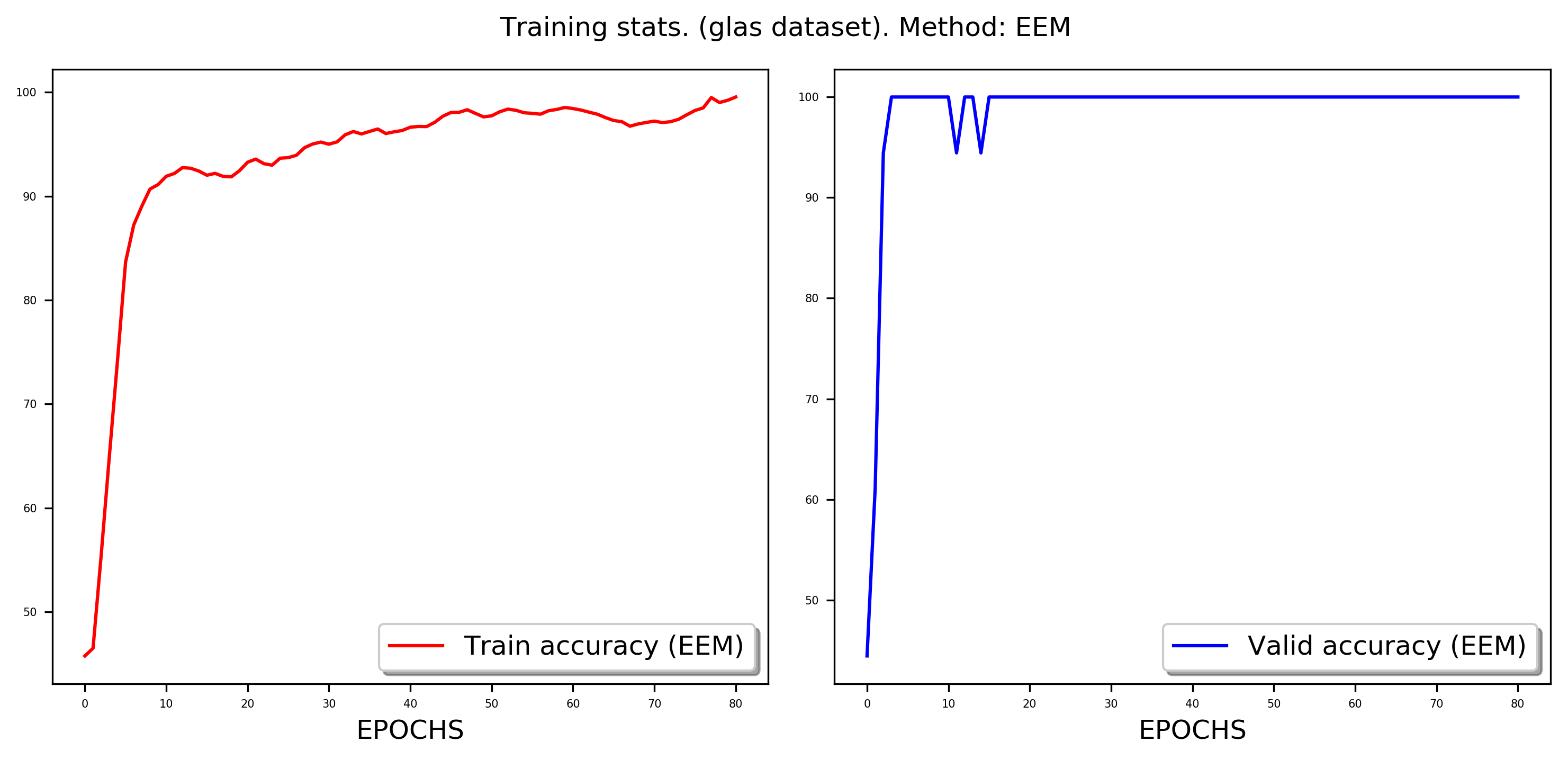}
\caption{Train and validation classification accuracy over GlaS dataset using EEM method.}
\label{fig:fig-glas-eem-train-curve}
\end{figure}

\begin{figure}[h!]
  \center
\includegraphics[width=1.\linewidth]{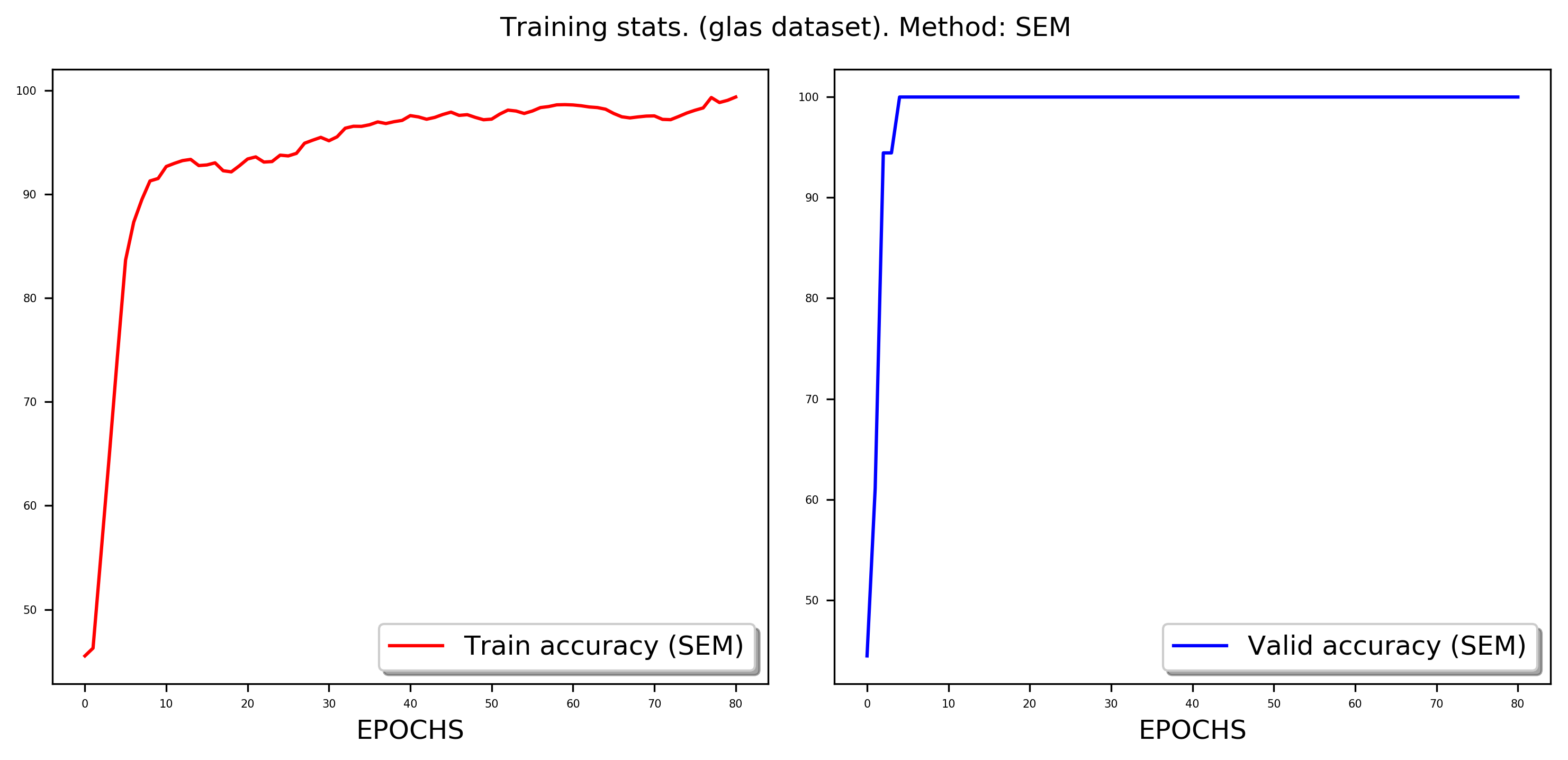}
\caption{Train and validation classification accuracy over GlaS dataset using SEM method.}
\label{fig:fig-glas-sem-train-curve}
\end{figure}

We report in Fig.\ref{fig:fig-glas-eem-train-curve} and \ref{fig:fig-glas-sem-train-curve} the training curves of EEM/SEM methods over the GlaS dataset. Despite their analytical difference, both methods lead to learning curves with quite similar shapes.



\bibliographystyle{apalike}
\bibliography{bibliography}

\end{document}